\pdfoutput=1

\documentclass[11pt]{article}

\usepackage{ACL2023}

\usepackage{times}
\usepackage{latexsym}
\usepackage{amsmath}
\usepackage{diagbox}
\usepackage{caption}
\usepackage{amsfonts,amssymb}
\usepackage{subfigure}
\usepackage{graphicx}
\usepackage{latexsym}
\usepackage{bm}
\usepackage{amssymb}
\usepackage{amsmath}
\usepackage{multirow}
\usepackage[ruled,lined,commentsnumbered]{algorithm2e}
\usepackage{hyperref}
\usepackage{booktabs}
\usepackage{bbm}
\usepackage{bm}
\usepackage[inkscapelatex=false]{svg}

\usepackage{amsmath}
\usepackage{verbatim}
\usepackage{booktabs}
\usepackage{bm}
\usepackage{amssymb}
\usepackage{pifont}
\usepackage{stfloats}
\usepackage{multicol}
\usepackage{float}
\usepackage{graphicx}
\usepackage{subfigure}
\usepackage{multirow}
\usepackage{array}
\newcommand{\PreserveBackslash}[1]{\let\temp=\\#1\let\\=\temp}
\newcolumntype{C}[1]{>{\PreserveBackslash\centering}p{#1}}
\newcolumntype{R}[1]{>{\PreserveBackslash\raggedleft}p{#1}}
\newcolumntype{L}[1]{>{\PreserveBackslash\raggedright}p{#1}}
\usepackage{diagbox}
\usepackage{mathrsfs}

\usepackage{mathtools}
\usepackage[T1]{fontenc}

\usepackage[utf8]{inputenc}

\usepackage{microtype}

\usepackage{inconsolata}

\title{Learning Optimal Policy for Simultaneous Machine Translation 

via Binary Search}

\author{
    Shoutao Guo \textsuperscript{\rm 1,2},
    Shaolei Zhang \textsuperscript{\rm 1,2},
    Yang Feng \textsuperscript{\rm 1,2}\thanks{ \ \ Corresponding author: Yang Feng.} \\
        \textsuperscript{\rm 1}{Key Laboratory of Intelligent Information Processing} \\ Institute of Computing Technology, Chinese Academy of Sciences (ICT/CAS) \\
    { \textsuperscript{\rm 2} {University of Chinese Academy of Sciences, Beijing, China}} \\
     \texttt{\{\href{mailto:guoshoutao22z@ict.ac.cn}{guoshoutao22z}, \href{mailto:zhangshaolei20z@ict.ac.cn}{zhangshaolei20z}, \href{mailto:fengyang@ict.ac.cn}{fengyang}\}@ict.ac.cn}  }

\begin{document}
\maketitle
\begin{abstract}
Simultaneous machine translation (SiMT) starts to output translation while reading the source sentence and needs a precise policy to decide when to output the generated translation. Therefore, the policy determines the number of source tokens read during the translation of each target token. However, it is difficult to learn a precise translation policy to achieve good latency-quality trade-offs, because there is no golden policy corresponding to parallel sentences as explicit supervision. In this paper, we present a new method for constructing the optimal policy online via binary search. By employing explicit supervision, our approach enables the SiMT model to learn the optimal policy, which can guide the model in completing the translation during inference. Experiments on four translation tasks show that our method can exceed strong baselines across all latency scenarios\footnote{Code is available at \url{https://github.com/ictnlp/BS-SiMT}}
\end{abstract}

\section{Introduction}

Simultaneous machine translation (SiMT) \citep{reinforcement, DBLP:conf/acl/MaHXZLZZHLLWW19, MIlk, DBLP:conf/iclr/MaPCPG20, MU}, which outputs the generated translation before reading the whole source sentence, is applicable to many real-time scenarios, such as live broadcast and real-time subtitles. To achieve the goal of high translation quality and low latency \citep{ITST}, the SiMT model relies on a policy that determines the number of source tokens to read during the translation of each target token.

The translation policy plays a pivotal role in determining the performance of SiMT, as an imprecise policy can lead to degraded translation quality or introduce unnecessary delays, resulting in poor translation performance \citep{DualPath}. Therefore, it is crucial to establish an optimal policy that achieves good latency-quality trade-offs. However, the absence of a golden policy between the source and target makes it challenging for the SiMT model to acquire the explicit supervision required for learning the optimal policy. According to \citet{MU}, the SiMT model will learn better policy if it is trained with external supervision. Consequently, by constructing the optimal policy between the source and target, we can train the SiMT model, which will then generate translations based on the learned policy during inference.

However, the existing methods, including fixed policy and adaptive policy, have limitations in learning the optimal policy due to the lack of appropriate explicit supervision. For fixed policy \citep{IncrementalDecoding, DBLP:conf/acl/MaHXZLZZHLLWW19, multiPath, DBLP:conf/emnlp/ZhangF21}, the model relies on heuristic rules to generate translations. However, these rules may not prompt the SiMT model to output the generated translation immediately, even when there is sufficient source information to translate the current target token. Consequently, the fixed policy often cannot achieve good latency-quality trade-offs because of its rigid rules. For adaptive policy \citep{reinforcement, MIlk, DBLP:conf/iclr/MaPCPG20, ITST}, the model can dynamically determine its policy based on the translation status, leading to improved performance. Nevertheless, precise policy learning without explicit supervision remains challenging. Some methods \citep{MU,translation-based} attempt to construct learning labels for the policy offline by introducing external information. But the constructed labels for policy learning cannot guarantee that they are also optimal for the translation model.

Under these grounds, our goal is to search for an optimal policy through self-learning during training, eliminating the need for external supervision.
Subsequently, this optimal policy can be employed to guide policy decisions during inference. In SiMT, increasing the number of source tokens read improves translation quality but also leads to higher latency \citep{DBLP:conf/acl/MaHXZLZZHLLWW19}. However, as the length of the read-in source sequence grows, the profit of translation quality brought by reading more source tokens will also hit bottlenecks \citep{DBLP:conf/emnlp/ZhangF21}. Therefore, the \emph{gain} of reading one source token can be evaluated with the ratio of the improvement in translation quality to the corresponding increase in latency. The optimal policy will make sure that every decision of reading or writing will get the greatest gain. In this way, after translating the whole source sequence, the SiMT can get the greatest gain, thereby achieving good latency-quality trade-offs.

In this paper, we propose a SiMT method based on binary search (BS-SiMT), which leverages binary search to construct the optimal translation policy online and then performs policy learning accordingly. Specifically, BS-SiMT model consists of a translation model and an agent responsible for policy decisions during inference. To construct the optimal policy, the translation model treats potential source positions as search interval and selects the next search interval by evaluating the concavity in binary search. This selection process effectively identifies the interval with the highest gain, thus enabling the construction of an optimal policy that ensures good performance. Subsequently, the constructed policy is used to train the agent, which determines whether the current source information is sufficient to translate the target token during inference. If the current source information is deemed sufficient, the translation model outputs the generated translation; otherwise, it waits for the required source tokens. Experiments on De$\leftrightarrow$En and En$\leftrightarrow$Vi translation tasks show that our method can exceed strong baselines under all latency.

\section{Background}
For SiMT task, the model incrementally reads the source sentence $\mathbf{x}$ = $(x_1, ... , x_J)$ with length $J$ and generates translation $\mathbf{y}$ = $(y_1, ... , y_I)$ with length $I$ according to a policy. To define the policy, we introduce the concept of the number of source tokens read when translating target token $y_i$, denoted as $g_i$. Then the translation policy can be formalized as $\mathbf{g}$ = $(g_1, ..., g_I)$. The probability of translating target token $y_i$ is $p_{\theta}(y_i | \mathbf{x}_{\leq g_i}, \mathbf{y}_{<i})$, where $\mathbf{x}_{\leq g_i}$ is the source tokens read in when translating $y_i$, $\mathbf{y}_{<i}$ is the output target tokens and $\theta$ is model parameters. Consequently, the SiMT model can be optimized by minimizing the cross-entropy loss:
\begin{equation}
\mathcal{L}_\textsc{ce} = - \sum\limits_{i = 1}^{I} \log p_{\theta}(y^{\star}_i | \mathbf{x}_{\leq g_i}, \mathbf{y}_{<i}),
\end{equation}
where $y^{\star}_i$ is the ground-truth target token. Because our policy is based on wait-$k$ policy \citep{DBLP:conf/acl/MaHXZLZZHLLWW19} and multi-path method \citep{multiPath}, we briefly introduce them.

\paragraph{Wait-$k$ policy}
For wait-$k$ policy \citep{DBLP:conf/acl/MaHXZLZZHLLWW19}, which is the most widely used fixed policy, the model initially reads $k$ source tokens and subsequently outputs and reads one token alternately. Therefore, $g_i$ is represented as:
\begin{equation}
    g^k_i = \min \{k+i-1, I \},
\end{equation}
where $I$ is the length of the source sentence.

\begin{figure}[t]
    \centering
    \includegraphics[width=3.0in]{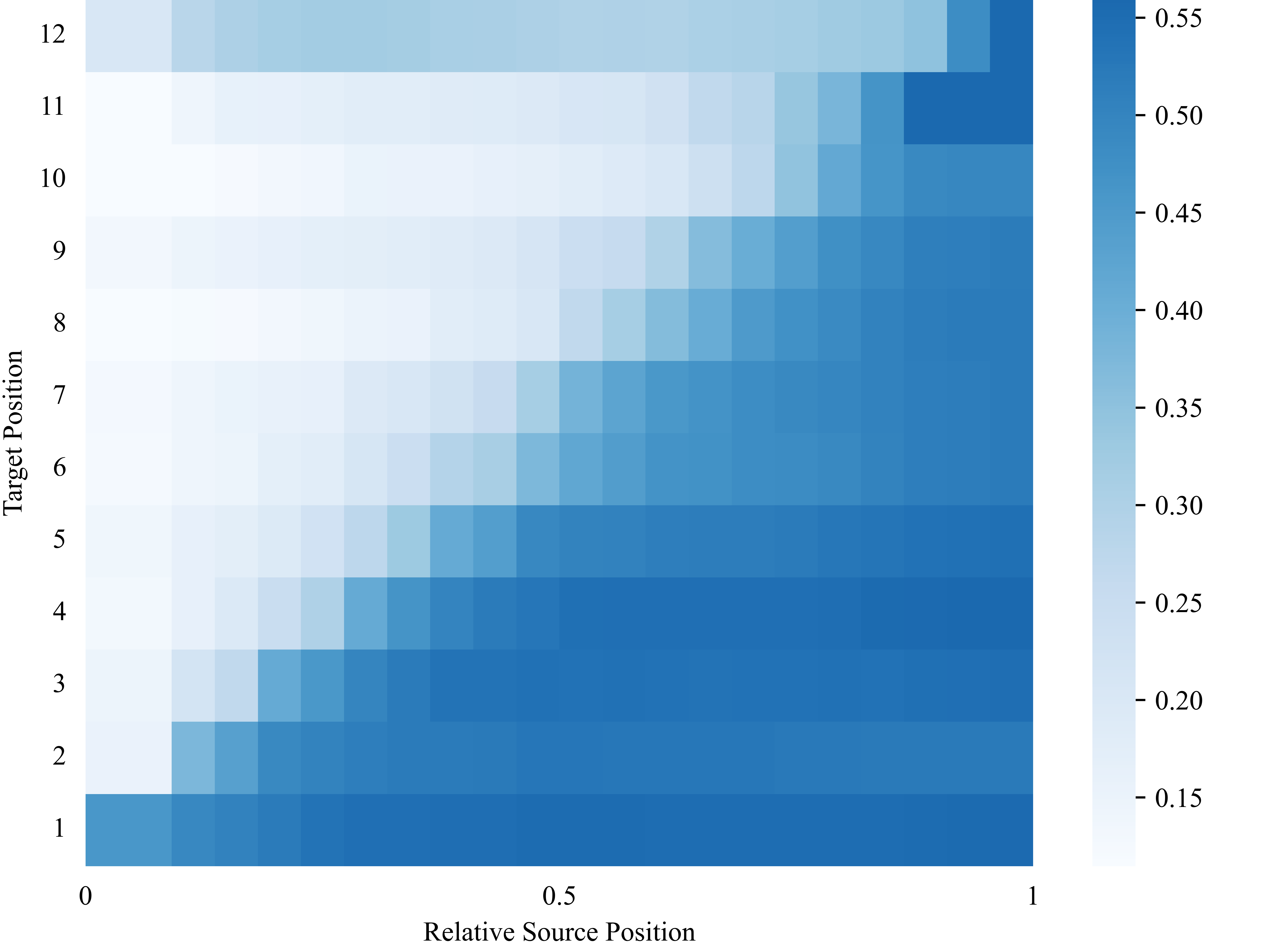}
    \caption{The translating probability of ground-truth when attending to different numbers of tokens. When translating each target token, the model adopts the wait-$k$ policy for previous tokens.
}
    \label{pre_alalysis}
\end{figure}

\paragraph{Multi-path}
To avoid the recalculation of the encoder hidden states every time a source token is read, multi-path \citep{multiPath} introduces a unidirectional encoder to make each source token only attend to preceding tokens. Furthermore, during training, the model can be trained under various by sampling latency $k$ uniformly:
\begin{equation}
\mathcal{L}_\textsc{ece} = - \sum\limits_{k \sim \mathcal{U}(\emph{\rm{K}})} \sum\limits_{i = 1}^{I} \log p_{\theta}(y^{\star}_i | \mathbf{x}_{\leq g^k_i}, \mathbf{y}_{<i}),
\end{equation}
where $k$ is uniformly sampled form \emph{\rm{K}} = $ [1,...,I]$. Therefore, the model can generate translation under all latency by only using a unified model.

\begin{figure*}[t]
    \centering
    \includegraphics[width=6.2in]{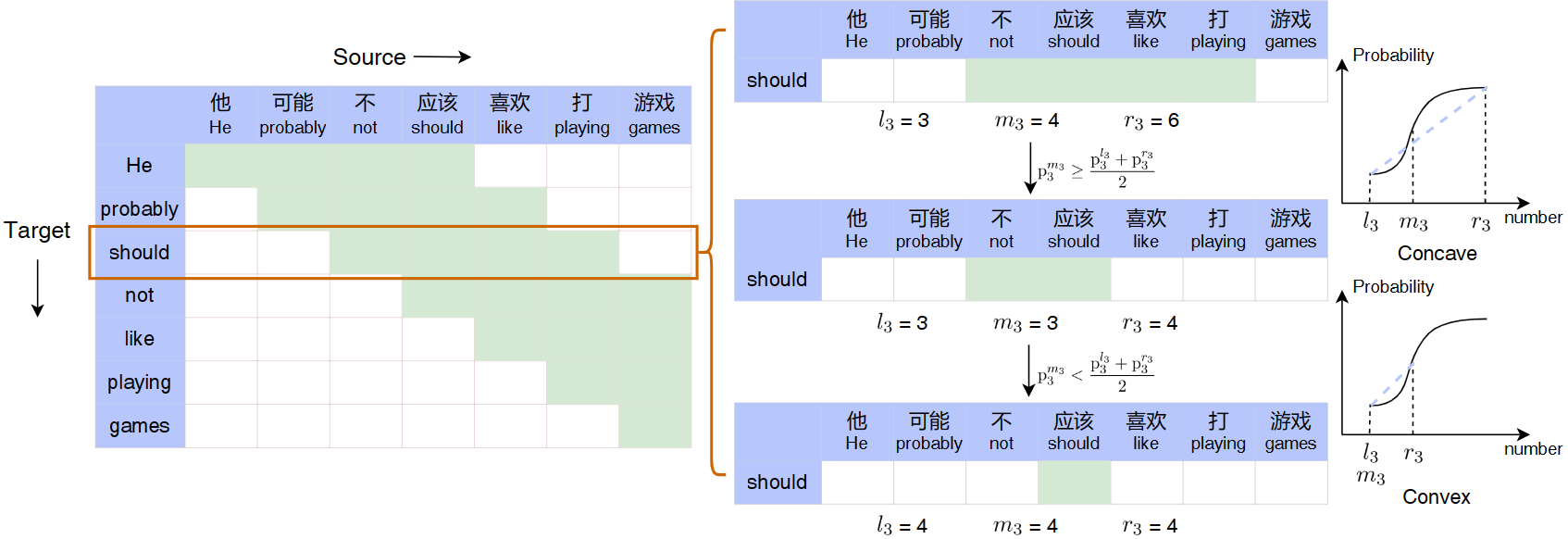}
    \caption{An example of finding the optimal policy through binary search. The light green area in the figure depicts the search interval of each target token. The horizontal axis of the two function images denotes the number of source tokens read in, and the vertical axis represents the probability of translating the ground-truth. Specifically, we focus on the search for the suitable number of source tokens required to translate the target token "should."}
    \label{fig-model}
\end{figure*}

\section{Preliminary Analysis}
In this section, we explore the influence of the number of read-in source tokens on translation quality. We employ the multi-path translation model \citep{multiPath} and select a bucket of samples from the IWSLT14 De$\rightarrow$En test set, consisting of 295 sentences with the same target length \citep{laf}. To analyze the variations, we utilize the probability of translating the ground-truth token as a measure of translation quality.
For each relative source position $q$, we compute the probability $p_{i}^{q}$ of translating the ground-truth $y_i^{\star}$:
\begin{equation}
     p_{i}^{q} = p(y_{i}^{\star} | \mathbf{x}_{\leq \lceil q*J \rceil}, \mathbf{y}_{<i}),
\end{equation}
where $J$ is the length of the source sentence, and compute the average $p_i^q$ across all samples. Since the lengths of the source sentences vary across different samples, we utilize the relative position, i.e., the proportion of the source position to the end of the sentence. The results in Figure \ref{pre_alalysis} show that the probability of translating target tokens increases with the number of source tokens. Notably, the necessary source tokens contribute the most to the improvement in translation quality. This finding suggests that translation quality often relies on the model obtaining the necessary source information, which is determined by the policy. This incremental nature observed here suggests that we can utilize binary search to get the policy, providing an important basis for our method.

\section{The Proposed Method}
\label{BS-SiMT}

Our BS-SiMT model contains two components: the translation model and the agent. The translation model, which is fine-tuned from the multi-path model, employs binary search to iteratively select the next interval with the highest gain. This process allows the model to search for the optimal policy and subsequently train itself based on the searched policy. Subsequently, we utilize the best-performing translation model to construct the optimal policy, which serves as explicit supervision for training the agent. During inference, the agent guides the translation model to generate translations with good latency-quality trade-offs. The details are introduced in the following sections.
\subsection{Constructing Optimal Policy}

The optimal policy ensures that the SiMT model gets good latency-quality trade-offs \citep{StreamLevelEvaluation}. The translation model plays a key role in searching for the optimal policy by identifying the number of source tokens to be read, maximizing the gain for the current translation. However, considering all possible numbers of source tokens for each target token would be computationally expensive and may not effectively balance latency and translation quality \citep{zhang2023hidden}. To address this issue, we employ binary search to determine the ideal number of source tokens to be read for each target token by evaluating the midpoint concavity of the interval.

To achieve this goal, we allocate the search interval of the number of source tokens for each target token. We denote the search interval for the target token $y_i$ as [$l_i, r_i$], where $l_i$ and $r_i$ represent the minimum and maximum number of source tokens to be considered, respectively. Then we can get the median value $m_i$ of the interval [$l_i, r_i$], which is calculated as:
\begin{equation}
     m_i = \lfloor \frac{l_i + r_i}{2} \rfloor.
\label{middle}
\end{equation}
Next, the probability $\mathrm{p}^{l_i}_i$ of translating ground-truth token $y^{\star}_i$ based on the previous $l_i$ source tokens can be calculated as follows:
\begin{equation}
    \mathrm{p}^{l_i}_i = p_{\theta}(y^{\star}_i | \mathbf{x}_{\leq l_i}, \mathbf{y}_{<i}).
\label{decProb}
\end{equation}
Similarly, $\mathrm{p}^{m_i}_i$ and $\mathrm{p}^{r_i}_i$ can also be calculated as Eq.(\ref{decProb}). We then discuss the conditions for selecting [$l_i, m_i$] or [$m_i\!+\!1, r_i$] as the next search interval. Obviously, the interval with a greater gain should be selected each time. The gain of interval [$l_i$, $m_i$] should be defined as:
\begin{equation}
    \frac{\mathrm{p}^{m_i}_i - \mathrm{p}^{l_i}_i}{m_i - l_i}.
\end{equation}

\begin{algorithm}[t]
  \SetAlgoLined
  \SetKwProg{Fn}{Function}{:}{end}
  \SetKwProg{ret}{return}{}{}
  \SetKwFunction{Eva}{Evaluation}
  \SetKwInput{KwData}{Data}
  \KwIn{Source sentence $\mathbf{x}$, Target sentence $\mathbf{y}$, Translation model $p_{\theta}()$}
  Initialize $l_i, r_i$
  
  \While{$l_i < r_i$}{
    calculate $m_i$ as Eq.(\ref{middle})
    
    calculate $\mathrm{p}^{l_i}_i, \mathrm{p}^{m_i}_i, \mathrm{p}^{r_i}_i$ as Eq.(\ref{decProb})
    
    \uIf{Eq.(\ref{condition}) is satisfied}{
        $r_i \leftarrow m_i$ \tcp*[f]{$\!\!\! $Left Range}
      }\uElse{
        $l_i \leftarrow m_i + 1$ \tcp*[f]{$\!\!\! $Right Range}
      }
    }
    $g_i = l_i$
  \caption{Search for Optimal Policy}
  \label{algor}
\end{algorithm}

Therefore, we select the interval with greater gain by comparing $\frac{\mathrm{p}^{m_i}_i - \mathrm{p}^{l_i}_i}{m_i - l_i}$ and $\frac{\mathrm{p}^{r_i}_i - \mathrm{p}^{m_i}_i}{r_i - m_i}$. Since $m_i - l_i$ is equal to $r_i - m_i$, it is actually a comparison between $\mathrm{p}^{m_i}_i$ and $\frac{\mathrm{p}^{l_i}_i + \mathrm{p}^{r_i}_i}{2}$. Hence, we select the interval [$l_i, m_i$] if the following condition is satisfied:
\begin{equation}
\mathrm{p}^{m_i}_i \geq \frac{\mathrm{p}^{l_i}_i + \mathrm{p}^{r_i}_i}{2}, 
\label{condition}
\end{equation}
otherwise we choose the interval [$m_i\!+\!1, r_i$]. The intuition behind this decision is that if the function composed of ($l_i, \mathrm{p}^{l_i}_i$), ($m_i, \mathrm{p}^{m_i}_i$), and ($r_i, \mathrm{p}^{r_i}_i$) exhibits midpoint concavity, we select the interval [$l_i, m_i$]; otherwise, we choose [$m_i\!+\!1, r_i$]. When the upper and lower boundaries of the search interval are the same, the model has found an appropriate policy. Figure \ref{fig-model} shows an example of finding the policy through binary search. We also provide a formal definition of the binary search process in Algorithm \ref{algor}. Importantly, the search process for all target tokens is performed in parallel.

The translation model undergoes iterative training to align with the searched policy, ensuring a gradual convergence. The optimization process of the translation model and the search for the optimal policy are carried out in an alternating manner. As a result, we construct the optimal translation policy $\mathbf{g}$ = $(g_1, ..., g_I)$ based on the search outcomes obtained from the best translation model. Besides, by adjusting the search interval, we can obtain the optimal translation policy under all latency.

\begin{figure}[t]
    \centering
    \includegraphics[width=2.5in]{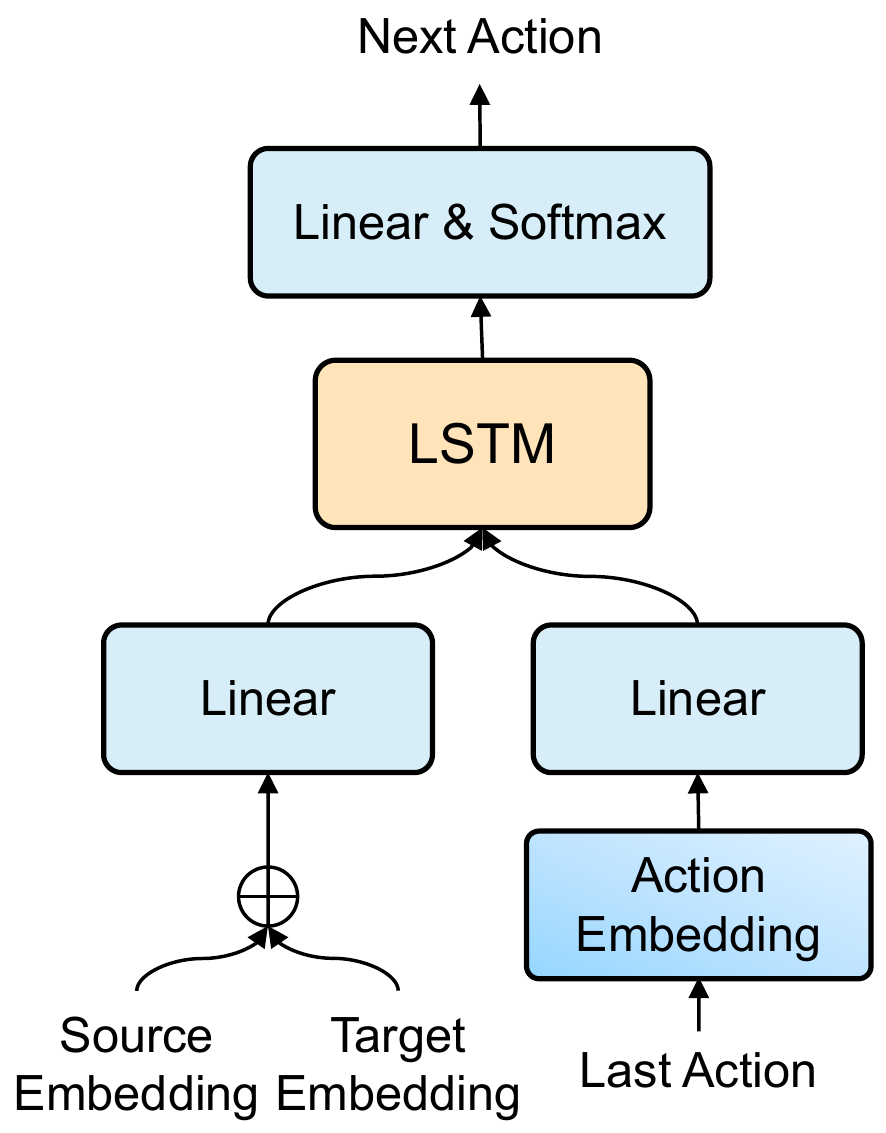}
    \caption{The architecture of the agent. The agent decides the next action based on the embedding of the last source and target token, as well as the last action.}
    \label{agent}
\end{figure}

\subsection{Learning Optimal Policy}
\label{learnOptimal}
Once the optimal translation policy is obtained for the corresponding parallel sentence, we can proceed to train the agent in order to learn this policy through explicit supervision. The agent will determine the policy based on the translation status during inference \citep{translation-based}. To facilitate this process, we introduce two actions: READ and WRITE. The READ action corresponds to reading the next source token, while the WRITE action represents outputting the generated translation. Instead of using the sequence $\mathbf{g}$ = $(g_1, ..., g_I)$ to represent the translation policy, we transform it into a sequence of READ and WRITE actions. This transformation is motivated by the fact that it is easier to determine the next action compared to predicting the number of source tokens required to translate the next target token based solely on the current translation status.

We denote the optimal action sequence as $\mathbf{a}$ = $(a_1, ... , a_T)$, where $T = I + J$. Consequently, the action to be taken at step $t$ can be derived from the optimal policy as follows:
\begin{equation}
    a_t = \left\{\begin{matrix}
  \text{WRITE},&\text{if} \;\; t=g_i + i\\
  \text{READ},& \text{otherwise}
  \end{matrix}\right. .
\end{equation}
The obtained optimal action sequence serves as the basis for training the agent to learn the optimal policy within a supervised framework. At step $t$, the agent receives the current translation status $o_t$, which includes the last source token $x_j$, the last generated token $y_i$, and the last action $a_{t-1}$. Based on this information, the agent determines the action $a_t$. We train the agent, implemented as an RNN architecture, to maximize the probability of the current action $a_t$ as follows:
\begin{equation}
    \max p_{{\theta}_a}(a_t | \mathbf{a}_{<t}, \mathbf{o}_{<t}),
\end{equation}
where ${\theta}_a$ is the parameters of the agent and $\mathbf{a}_{<t}$, and $\mathbf{o}_{<t}$ represent the sequence of actions and the translation status before time step $t$, respectively.

The architecture of the agent is shown in Figure \ref{agent}. At each step, the agent receives the embedding of the last source and target token,  along with the last action. The embedding of the last source and target token, generated by the translation model, is concatenated and passed through a linear layer. The last action is also processed through a separate embedding and linear layer. Subsequently, the outputs of the two linear layers will be fed into an LSTM layer \citep{DBLP:journals/neco/HochreiterS97} to predict the next action. Furthermore, to mitigate the mismatch between training and testing, we train the agent using the embeddings of the generated translation instead of relying on the ground-truth.

\begin{algorithm}[t]
  \SetAlgoLined
  \SetKwProg{Fn}{Function}{:}{end}
  \SetKwProg{ret}{return}{}{}
  \SetKwFunction{Eva}{Evaluation}
  \SetKwInput{KwData}{Data}
  \KwIn{Source sentence $\mathbf{x}$, Translation model $p_{\theta}()$, Agent $p_{{\theta}_a}()$}
  $y_0 \leftarrow \left \langle bos \right \rangle$, $a_1 \leftarrow \;$READ
  
  $i \leftarrow 1$, $j \leftarrow 1$, $t \leftarrow 2$
  
  \While{$y_{i-1} \neq \left \langle eos \right \rangle$}{
    decide $a_t$ using translation status
    
    \uIf{$a_t = \;$\rm{WRITE} \bf{or} $x_j = \left \langle eos \right \rangle$}{
        generate $y_i$
        
        $i \leftarrow i + 1$
      }\uElse{
        read the next token
        
        $j \leftarrow j + 1$
      }
      $t \leftarrow t+1$
    }
  \caption{The Process of Inference}
  \label{algor-infer}
\end{algorithm}

\subsection{Inference}
Up to now, we get the trained translation model and agent. Our BS-SiMT model generates translations by leveraging the translation model, which is guided by the agent for policy decisions. At each step, the agent receives the translation status from the translation model and determines the next action. Then the translation model either outputs translation or reads the next source token based on the decision of the agent. The inference process is formally expressed in Algorithm \ref{algor-infer}.

\begin{figure*}[t]
\centering
\subfigure[En$\rightarrow $Vi]{
\includegraphics[width=1.55in]{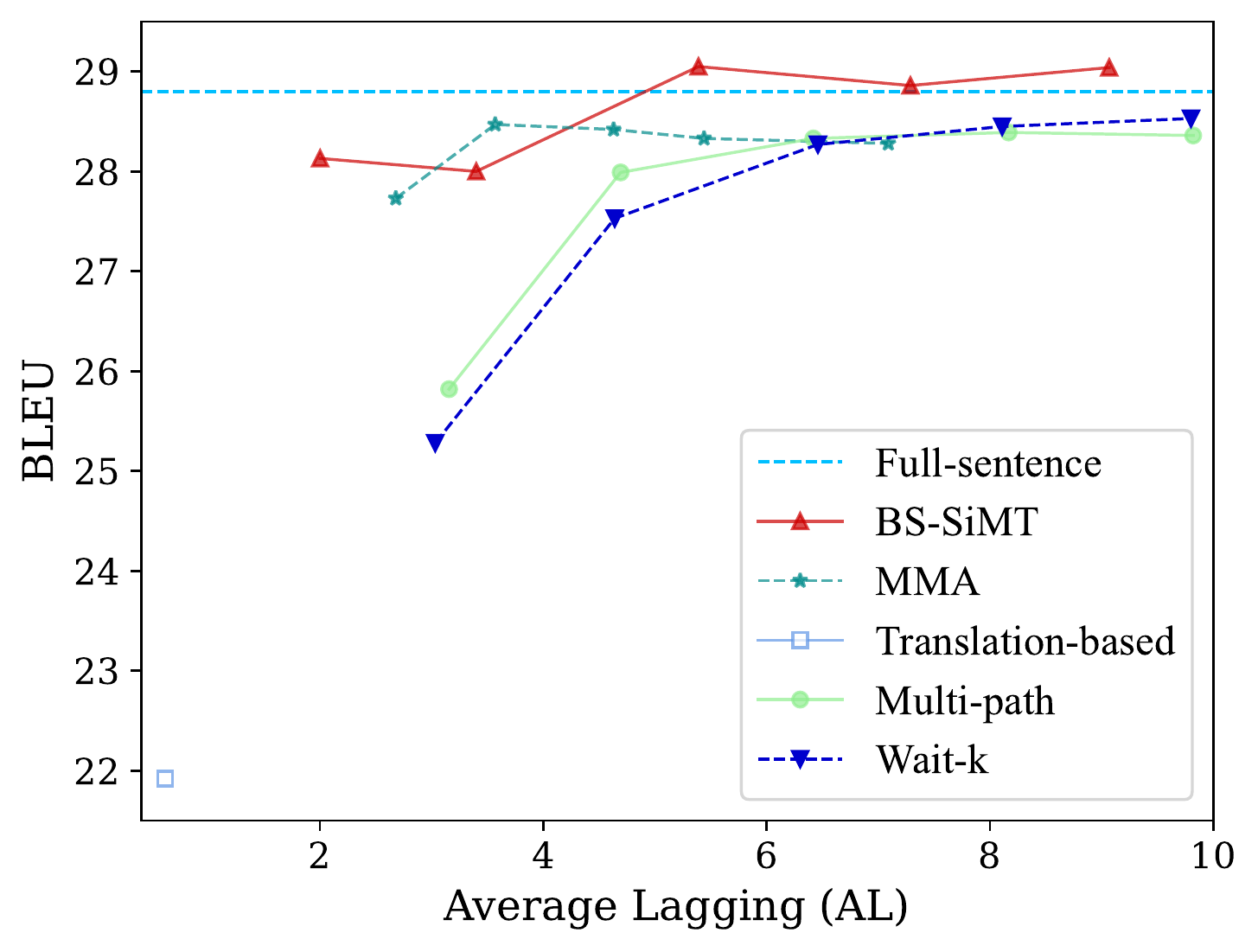}
}\hspace{-0.3cm}
\subfigure[Vi$\rightarrow $En]{
\includegraphics[width=1.55in]{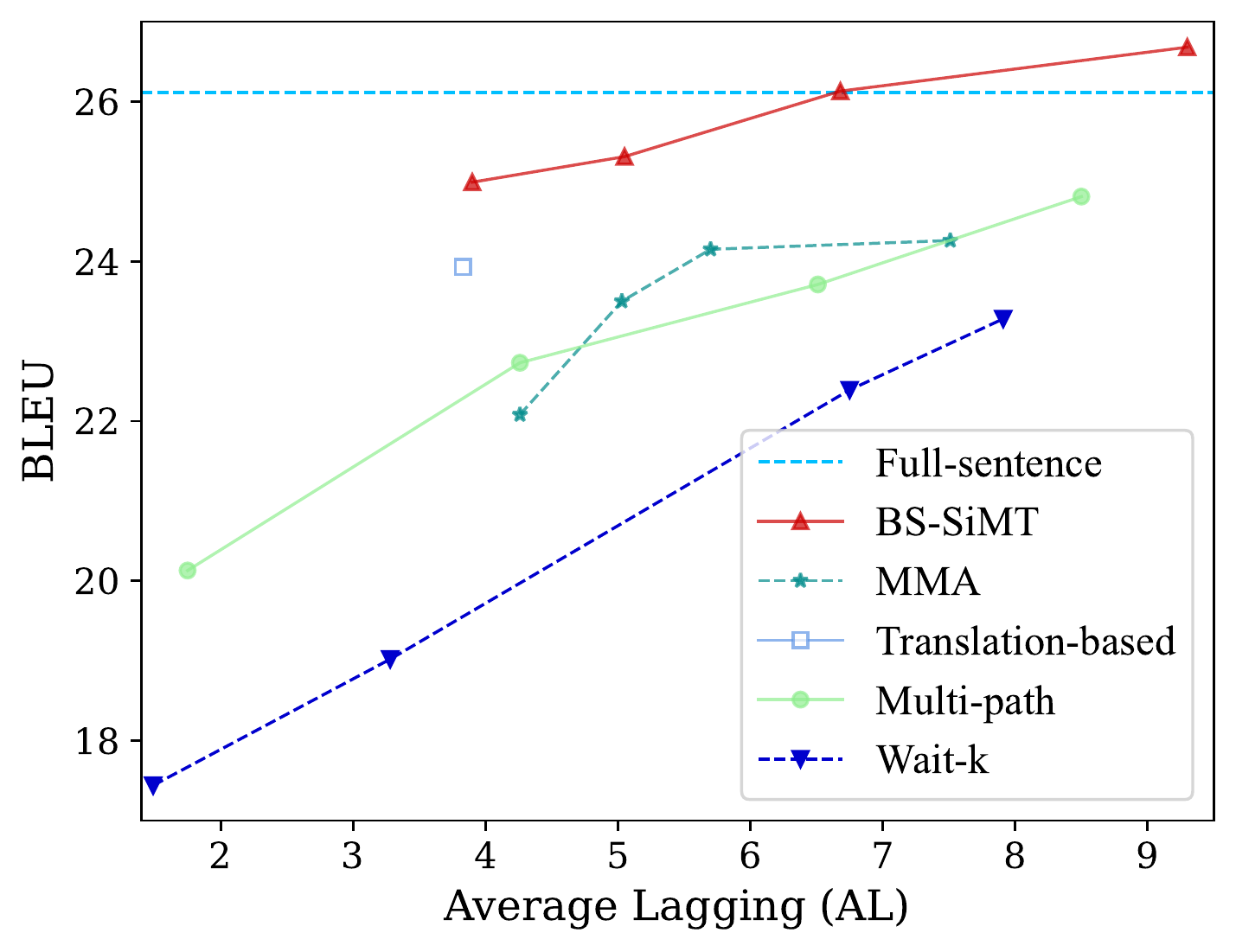}
}\hspace{-0.3cm}
\subfigure[De$\rightarrow $En]{
\includegraphics[width=1.55in]{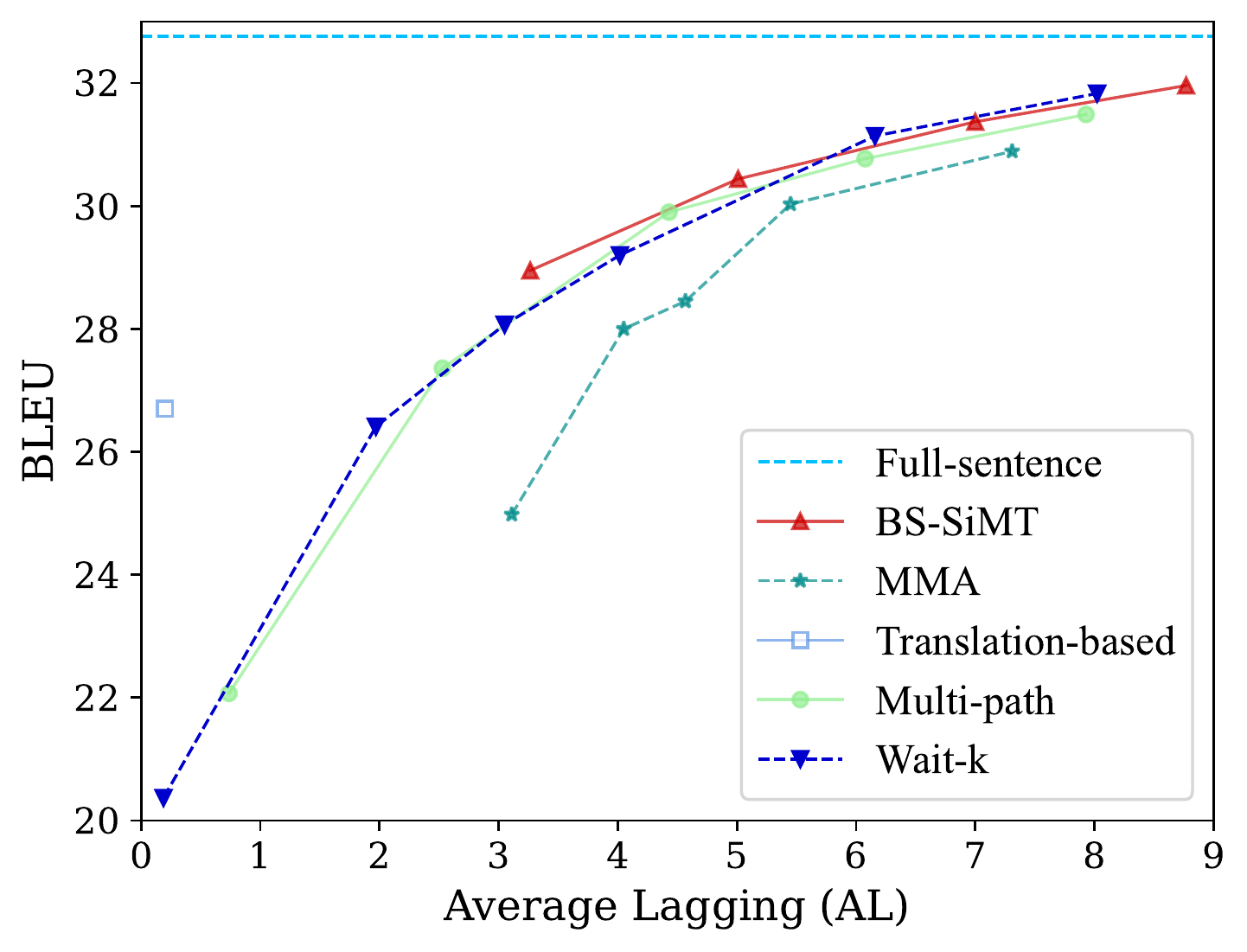}
}\hspace{-0.3cm}
\subfigure[En$\rightarrow $De]{
\includegraphics[width=1.55in]{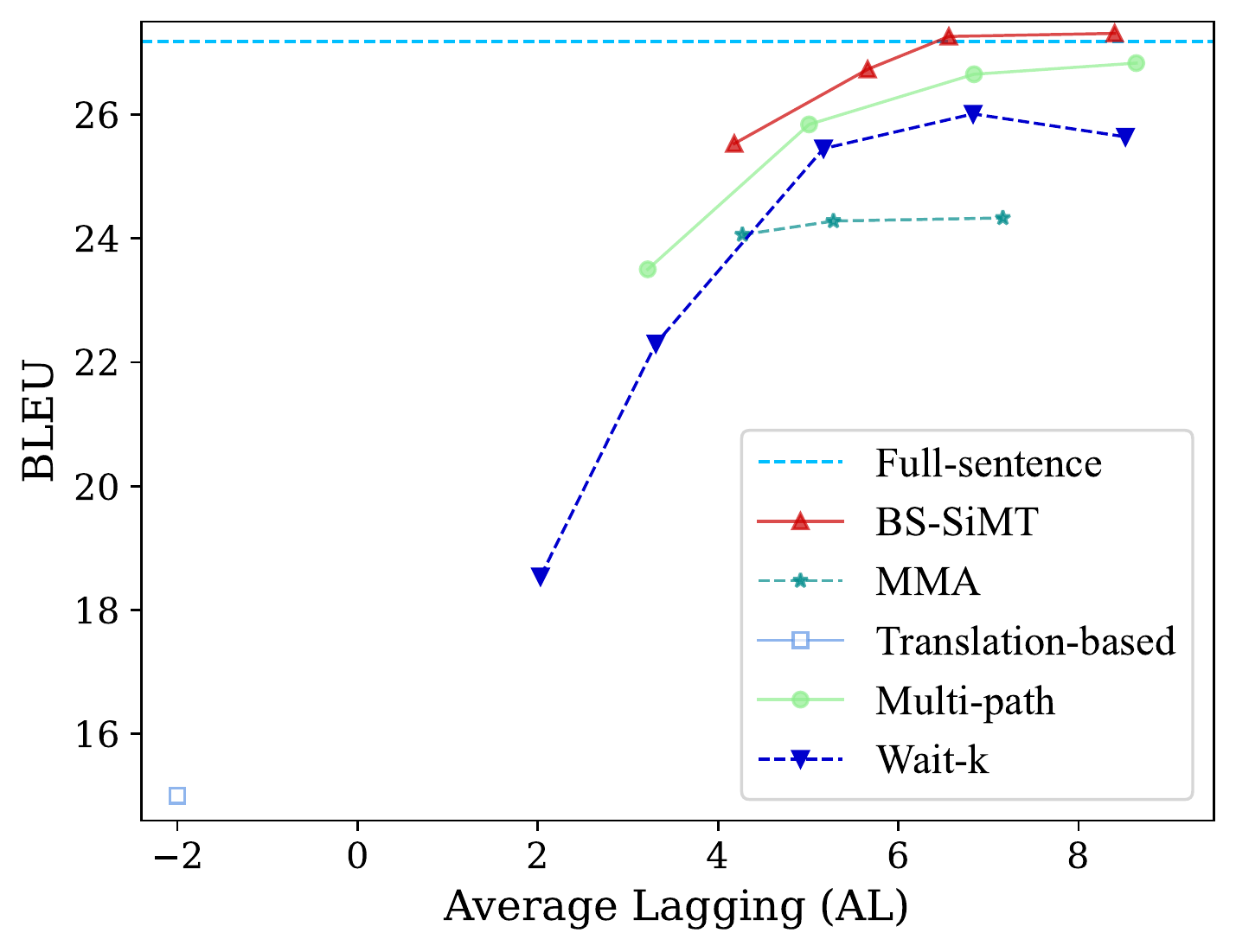}
}

\caption{Translation performance of different methods on En$\leftrightarrow$Vi and De$\leftrightarrow$En tasks. It shows the results of our BS-SiMT method, Wait-$k$ policy, Multi-path, Translation-based policy, MMA policy, and the Full-sentence translation model.}
\label{main}
\end{figure*}

\section{Experiments}
\subsection{Datasets}
We evaluate our BS-SiMT method mainly on IWSLT15\footnote{\url{https://nlp.stanford.edu/projects/nmt/}} English$\leftrightarrow$Vietnamese (En$\leftrightarrow$Vi) and IWSLT14\footnote{\url{https://wit3.fbk.eu/2014-01}} German$\leftrightarrow$English (De$\leftrightarrow$En) tasks.

For En$\leftrightarrow$Vi task \citep{DBLP:conf/iwslt/CettoloNSBCF16}, our settings are the same as \citet{MIlk}. We use TED tst2012 as the development set and TED tst2013 as the test set. We replace tokens whose frequency is less than 5 with $\left \langle unk \right \rangle$.

For De$\leftrightarrow$En task, we keep our settings consistent with \citet{translation-based}. We use a concatenation of dev2010 and tst2010 to tst2013 as the test set. We apply BPE \citep{BPE} with 10K merge operations, which results in 8.8K German and 6.6K English sub-word units.

\subsection{Model Settings}
Since our experiments involve the following methods, we briefly introduce them.

{\bf {Wait-}$k$} Wait-$k$ policy \citep{DBLP:conf/acl/MaHXZLZZHLLWW19} reads $k$ source tokens first and then writes a target token and reads a source token alternately.

{\bf {Multi-path}} Multi-path \citep{multiPath} introduces a unidirectional encoder and trains the model by uniformly sampling the latency.


{\bf {MMA}} MMA \citep{DBLP:conf/iclr/MaPCPG20}, which is a superior adaptive policy in SiMT, allows each head to decide the policy independently and integrates the results of multiple heads.

{\bf {Translation-based }} Translation-based policy \citep{translation-based} decides its policy by comparing the translation of the Full-sentence translation model with the results of other policies.

\begin{table}[]
\centering
\begin{tabular}{c c|C{1.2cm}C{1.2cm}} \toprule[1.2pt]
\textbf{Length} & \textbf{[$l_1, r_1$]} & \textbf{AL} & \textbf{BLEU} \\ \midrule[0.8pt]
\multirow{2}{*}{5} & [3, 7] & \textbf{3.26} & \textbf{28.95}     \\
        & [5, 9] & \textbf{5.01} & \textbf{30.44} \\ \cline{1-4}

\multirow{2}{*}{3} & [3, 5] & 3.22 & 28.29     \\
        & [5, 7] & 5.88 & 30.69 \\ \cline{1-4}
\multirow{2}{*}{7} & [3, 9] & 3.94 & 26.76     \\
        & [5, 11] & 5.41 & 29.14 \\ 
\bottomrule[1pt]
\end{tabular}
\caption{Comparison of different lengths of search interval. The hyperparameter is the search interval for the first target token and the search interval for subsequent target tokens shifts one unit to the right from the previous one.}
\label{DiffInterval}
\end{table}

{\bf {Full-sentence}} Full-sentence is the conventional full-sentence translation model based on Transformer \citep{Transformer}.

{\bf {BS-SiMT}} Our proposed method in section \ref{BS-SiMT}.

The implementations of all our methods are adapted from Fairseq Library \citep{DBLP:conf/naacl/OttEBFGNGA19}, which is based on Transformer \citep{Transformer}. We apply the Transformer-Small model with 6 layers and 4 heads to all translation tasks. For Translation-based policy and our BS-SiMT, we augment the implementation by introducing the agent to make decisions for actions. The translation model of our BS-SiMT is fine-tuned from Multi-path. For our method, we set the model hyperparameter as the search interval [$l_1, r_1$] for the first target token, and the search interval for subsequent target tokens is shifted one unit to the right from the previous token. The agent is composed of 1-layer LSTM \citep{DBLP:journals/neco/HochreiterS97} with 512 units, 512-dimensional embedding layers, and 512-dimensional linear layers. Other model settings follow \citet{DBLP:conf/iclr/MaPCPG20}. We use greedy search at inference and evaluate these methods with translation quality measured by tokenized BLEU \citep{BLEU} and latency estimated by Average Lagging (AL) \citep{DBLP:conf/acl/MaHXZLZZHLLWW19}. 

\begin{table}[]
\centering
\begin{tabular}{c c|C{1.2cm}C{1.2cm}} \toprule[1.2pt]
\textbf{Reference} & \textbf{[$l_1, r_1$]} & \textbf{AL} & \textbf{BLEU} \\ \midrule[0.8pt]
\multirow{2}{*}{Translation} & [3, 7] & \textbf{3.26} & \textbf{28.95}     \\
        & [5, 9] & \textbf{5.01} & \textbf{30.44} \\ \cline{1-4}

\multirow{2}{*}{Ground-Truth} & [3, 7] & 3.24 & 28.41     \\
        & [5, 9] & 5.20 & 30.19 \\
\bottomrule[1pt]
\end{tabular}
\caption{Model performance of different translation status during training of the agent.}
\label{DiffReference}
\end{table}

\subsection{Main Results}
The translation performance comparison between our method and other methods on 4 translation tasks is shown in Figure \ref{main}. Our BS-SiMT method consistently outperforms the previous methods under all latency and even exceeds the performance of the Full-sentence translation model with lower latency on En$\rightarrow$Vi, Vi$\rightarrow$En, and En$\rightarrow$De tasks. This shows the effectiveness of our method.

\begin{table*}[]
\centering
\begin{tabular}{c|C{1.2cm}C{1.2cm}C{1.2cm}C{1.2cm}|C{1.2cm}C{1.2cm}C{1.2cm}C{1.2cm}} \toprule[1.2pt]
\textbf{Method} & \multicolumn{4}{c|}{\textbf{BS-SiMT}} & \multicolumn{4}{c}{\textbf{Oracle Policy}}                                      \\\cline{2-9}
\textbf{[$l_1, r_1$]}    & \textbf{[3, 7]} & \textbf{[5, 9]} & \textbf{[7, 11]} & \textbf{[9, 13]}  & \textbf{[3, 7]} & \textbf{[5, 9]} & \textbf{[7, 11]} & \textbf{[9, 13]} \\ \midrule[0.8pt]
\textbf{AL}        & 3.26 & 5.01 & 7.00 & 8.77 & 3.27 & 5.29 & 7.19 & 8.95 \\
\textbf{BLEU}      & 28.95 & 30.44 & 31.37 & 31.96 & 29.67 & 30.82 & 31.50 & 31.99 \\ 
\bottomrule[1pt]
\end{tabular}
\caption{Comparison between BS-SiMT (the trained agent) and Oracle Policy. We change the search interval for the first target token to achieve translation under all latency. }
\label{oraclePolicy}
\end{table*}

Compared to Wait-$k$ policy, our method obtains significant improvement. This improvement can be attributed to the dynamic policy decision in our method, where the policy is based on the translation status.  In contrast,  Wait-$k$ policy relies on heuristic rules for translation generation. Our method also surpasses Multi-path method greatly since it only changes the training method of the translation model, but still performs fixed policy during inference \citep{multiPath}. Compared to MMA, which is the superior policy in SiMT, our method achieves comparable performance and demonstrates better stability under high latency. MMA allows each head to independently decide its policy and perform translation concurrently, which can be affected by outlier heads and impact overall translation performance, particularly under high latency \citep{DBLP:conf/iclr/MaPCPG20}. In contrast, our method separates the policy and translation model, resulting in improved stability and efficiency \citep{MU}. When compared to the Translation-based policy, our method outperforms it and is capable of generating translation under all latency. Translation-based policy, which obtains the labels by utilizing external translation of the Full-sentence model, can only obtain the translation under a certain latency because of its offline construction method \citep{translation-based}. In contrast, our method constructs the optimal policy online while taking into account the performance of the translation model, thereby getting better latency-quality trade-offs. Additionally, our method surpasses the Full-sentence model on En$\rightarrow$Vi, Vi$\rightarrow$En, and En$\rightarrow$De tasks, highlighting the critical role of the policy in SiMT performance.

\begin{figure}[t]
    \centering
    \includegraphics[width=\columnwidth]{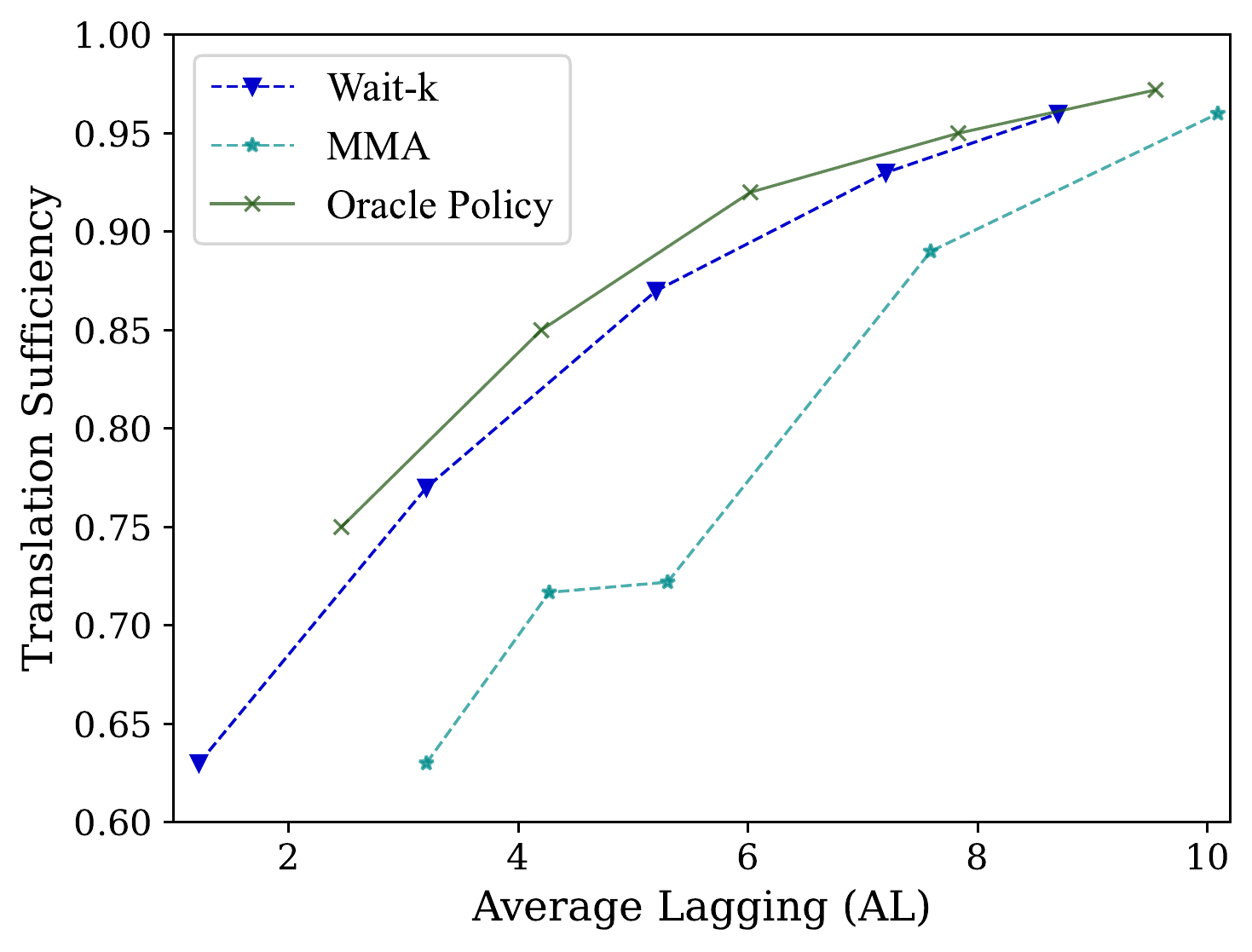}
    \caption{Comparison of translation sufficiency of different translation policies.}
    \label{sufficiency}
\end{figure}

\section{Analysis}
To gain insights into the improvements achieved by our method, we conduct extensive analyses. All of the following results are reported on De$\rightarrow$En task. The results presented below provide a detailed understanding of our method.

\begin{table}[]
\centering
\begin{tabular}{c c|C{1.2cm}C{1.2cm}} \toprule[1.2pt]
\textbf{Method} & \textbf{[$l_1, r_1$]} & \textbf{AL} & \textbf{BLEU} \\ \midrule[0.8pt]
\multirow{2}{*}{Concavity} & [3, 7] & \textbf{3.26} & \textbf{28.95}     \\
        & [5, 9] & \textbf{5.01} & \textbf{30.44} \\ \cline{1-4}

\multirow{2}{*}{GT} & [3, 7] & 4.81 & 20.85     \\
        & [5, 9] & 6.61 & 22.81 \\

\bottomrule[1pt]
\end{tabular}
\caption{Comparison of different trade-off approaches. `Concavity' indicates building optimal policy by checking concavity. `GT' indicates building optimal policy by comparing translation and ground-truth.}
\label{Trade-off}
\end{table}

\subsection{Ablation Study}
We conducted ablation studies to investigate the impact of the search interval and translation status on our BS-SiMT model. Regarding the search interval, we explore the effect of different lengths of search interval on translation performance. As shown in Table \ref{DiffInterval}, our BS-SiMT model, with a search interval of 5, surpasses other settings. This finding highlights the effectiveness of setting an appropriate search interval close to the diagonal for each target token \citep{zhang2023hidden}. By adjusting the search interval of the target tokens, we can obtain the optimal policy under all latency.

Additionally, we explored the influence of the translation status on the agent. As mentioned in subsection \ref{learnOptimal}, the agent determines its action based on the current translation status, which includes the last generated token. Hence, it is crucial to investigate whether using the generated translation or ground-truth in training the agent yields better results. As shown in Table \ref{DiffReference}, the agent trained with generated translation demonstrates superior performance. This can be attributed to the deviation between the ground-truth and the translation status obtained by the model during inference. Training the agent with the generated translation enables a better alignment between its training and testing conditions, resulting in improved performance.

\begin{table}[]
\centering
\begin{tabular}{c c|C{1.2cm}C{1.2cm}} \toprule[1.2pt]
\textbf{Base Model} & \textbf{[$l_1, r_1$]} & \textbf{AL} & \textbf{BLEU} \\ \midrule[0.8pt]
\multirow{2}{*}{Multi-path} & [3, 7] & \textbf{3.26} & \textbf{28.95}     \\
        & [5, 9] & \textbf{5.01} & \textbf{30.44} \\ \cline{1-4}

\multirow{2}{*}{Full-sentence} & [3, 7] & 3.83 & 28.80     \\
        & [5, 9] & 5.59 & 30.28 \\ \cline{1-4}

\multirow{2}{*}{None} & [3, 7] & 3.43 & 26.90     \\
        & [5, 9] & 5.25 & 28.46 \\
\bottomrule[1pt]
\end{tabular}
\caption{Comparison of different training methods of translation model. `Full-sentence' indicates the translation model is fine-tuned from the Full-sentence translation model. `None' represents the translation model is trained from scratch.}
\label{DiffBaseModel}
\end{table}

\subsection{Performance of Oracle Policy}
In addition to the ablation study, we also compare the performance on the test set according to the oracle policy. The oracle policy is obtained by our translation model using the whole source sentence on the test set. Therefore, the oracle policy is actually the optimal policy obtained by our method on the test set. As shown in Table \ref{oraclePolicy}, our oracle policy can achieve high translation quality, especially under low latency. This reflects the effectiveness of our way of building the optimal policy and our learned policy still has room for improvement.

A good policy needs to ensure that the target token is generated only after the required source information is read. To evaluate the constructed oracle policy, we introduce sufficiency \citep{DualPath} as the evaluation metric. Sufficiency measures whether the number of source tokens read exceeds the aligned source position when translating each target token, thus reflecting the faithfulness of the translation.

We evaluate the sufficiency of translation policy on RWTH De$\rightarrow$En alignment dataset\footnote{\url{https://www-i6.informatik.rwth-aachen.de/goldAlignment/}}, where reference alignments are annotated by experts and seen as golden alignments\footnote{For one-to-many alignment from target to source, we choose the position of farthest aligned source token.}. The results are shown in Figure \ref{sufficiency}. The oracle policy performs better than other methods in sufficiency evaluation and can even cover 75$\%$ of the aligned source tokens under low latency. Wait-$k$ policy is worse than our oracle policy under low latency because it may be forced to output translation before reading the aligned source tokens \citep{DBLP:conf/acl/MaHXZLZZHLLWW19}. MMA gets the worst performance in sufficiency evaluation, which may be attributed to its serious problem of outlier heads on De$\rightarrow$En task. Combined with the results in Figure \ref{main}, our oracle policy achieves good trade-offs by avoiding unnecessary latency while ensuring translation faithfulness.

\subsection{Analysis of the Trade-off Approach}

Our BS-SiMT approach achieves trade-offs by evaluating the concavity during binary search and selecting the interval with greater gain. Whether this trade-off approach is better needs to be further explored. In our method, we also consider an alternative approach within the framework. We investigate whether comparing the translation and ground-truth can be used to construct the optimal policy. As shown in Table \ref{Trade-off}, our method performs better than comparing translation and ground-truth. This is mainly because the condition of the latter method is difficult to achieve, resulting in the model reading too many source tokens \citep{MU}. Our approach allows for a broader interval to obtain translation policy, enabling the construction of a more effective translation policy.

\begin{table}[]
\centering
\begin{tabular}{c c|C{1.2cm}C{1.2cm}} \toprule[1.2pt]
\textbf{Architecture} & \textbf{[$l_1$, $r_1$]} & \textbf{AL} & \textbf{BLEU} \\ \midrule[0.8pt]
\multirow{2}{*}{LSTM} & [3, 7] & \textbf{3.26} & \textbf{28.95}     \\
        & [5, 9] & \textbf{5.01} & \textbf{30.44} \\ \cline{1-4}

\multirow{2}{*}{GRU} & [3, 7] & 3.34 & 28.19     \\
        & [5, 9] & 5.18 & 30.43 \\ \cline{1-4}

\multirow{2}{*}{Linear} & [3, 7] & 3.65 & 27.82     \\
        & [5, 9] & 5.60 & 29.99 \\
\bottomrule[1pt]
\end{tabular}
\caption{Performance comparison of different architectures of the agent. `GRU' and `Linear’ represent that the agent adopts GRU and Linear architecture respectively.}
\label{Diffagent}
\end{table}

\subsection{Training of Translation Model}
In our method, the construction of the optimal policy relies on the performance of the translation model. Therefore, the training of the translation model needs to be further explored. As shown in Table \ref{DiffBaseModel}, our method obtains the best performance. Training from scratch yields the worst performance, as the model lacks the ability to distinguish between good and poor translations. Fine-tuning from the Full-sentence model achieves better performance, but it does not have the ability to generate high-quality translation with partial source information. Our method, fine-tuned from Multi-path, is capable of generating high-quality translation under all latency.

\subsection{Analysis on the Trained Agent}
As introduced in subsection \ref{learnOptimal}, the agent is trained with the constructed optimal policy. The training of the agent becomes a supervised learning process. Thus, we need to analyze the impact of different architectures of the agent on our method. The results presented in Table \ref{Diffagent} demonstrate that the LSTM architecture achieves the best performance. On the other hand, the linear model with one hidden layer performs the worst due to its limited capacity to model sequential information compared to the RNN architecture. The LSTM model, with its larger number of trainable parameters, proves to be more suitable for this task than the GRU model.

\section{Related Work}
Recent SiMT methods can be roughly divided into two categories: fixed policy and adaptive policy.

For fixed policy, the model relies on predefined heuristic rules to generate translations. \citet{IncrementalDecoding} proposed STATIC-RW, which reads and writes \rm{RW} tokens alternately after reading $S$ tokens. \citet{DBLP:conf/acl/MaHXZLZZHLLWW19} proposed Wait-$k$ policy, which writes and reads a token alternately after reading $k$ tokens. \citet{multiPath} introduced the unidirectional encoder and enhanced Wait-$k$ policy by uniformly sampling latency $k$ during training. \citet{DBLP:conf/aaai/ZhangFL21} proposed future-guided training to help SiMT model invisibly embed future source information through knowledge distillation. \citet{char-waitk} proposed char-level Wait-$k$ policy to make the SiMT model adapt to the streaming input environment. \citet{DBLP:conf/emnlp/ZhangF21} proposed MoE wait-$k$ policy, which makes different heads execute different Wait-$k$ policies, and combine the results under multiple latency settings to predict the target tokens.

For adaptive policy, the translation policy is determined based on current translation status. \citet{reinforcement} trained the agent for policy decisions using reinforcement learning. \citet{SimplerAndFaster} trained the agent with optimal action sequences generated by heuristic rules. \citet{MIlk} proposed MILk, which applies the monotonic attention and determines the policy based on a Bernoulli variable. \citet{DBLP:conf/iclr/MaPCPG20} proposed MMA, which implements MILk on Transformer architecture and achieves superior performance in SiMT. \citet{MU} proposed MU, which is an adaptive segmentation policy \citep{diseg}. \citet{translation-based} used a full-sentence model to construct the translation policy offline, which can be used to train the agent. \citet{gaussian} implemented the adaptive policy by predicting the aligned source positions of each target token directly. \citet{DualPath} introduced dual constraints to make forward and backward models provide path supervision for each other. \citet{DBLP:conf/emnlp/ZhangGF22} proposed the Wait-info policy to balance source and target at the information level. \citet{post-eval} performed the adaptive policy by integrating post-evaluation into the fixed policy. \citet{zhang2023hidden} proposed Hidden Markov Transformer, which models simultaneous machine translation as a hidden Markov process.

The previous methods often lack explicit supervision for the learning of the policy. Some papers use external information, such as generated heuristic sequences, to learn the policy \citep{SimplerAndFaster, MU, translation-based}. However, their methods heavily rely on heuristic rules and offline reference sequence construction, which affects the translation performance. Our BS-SiMT constructs the optimal translation policy online by checking the concavity via binary search without utilizing external information, thereby obtaining good latency-quality trade-offs.

\section{Conclusion}
In this paper, we propose BS-SiMT, which utilizes binary search to construct the optimal translation policy online, providing explicit supervision for the agent to learn the optimal policy. The learned policy effectively guides the translation model in generating translations during inference. Experiments and extensive analyses show that our method can exceed strong baselines under all latency and learn a translation policy with good trade-offs.

\section*{Limitations}
In this paper, we build the optimal translation policy under all latency by simply setting the search interval, achieving high performance. However, we think that the performance of our method can be further improved by exploring more interval settings. Additionally, although we train the agent using a simple architecture and achieve good performance, there exists a performance gap between the learned policy and the searched optimal policy under low latency. Exploring more powerful models of the agent may help improve the performance and we leave it for future work.

\section*{Acknowledgment}

We thank all anonymous reviewers for their valuable suggestions. This work was supported by the National Key R$\&$D Program of China (NO. 2018AAA0102502).

\bibliography{anthology,custom}
\bibliographystyle{acl_natbib}

\appendix

\section{Hyperparameters}
All system settings in our experiments are shown in Table \ref{Hyper}.

\section{Numerical Results}
Table \ref{envi}, \ref{vien}, \ref{deen}, \ref{ende} respectively report the numerical results on IWSLT15 En$\rightarrow $Vi, IWSLT15 Vi$\rightarrow $En, IWSLT14 De$\rightarrow $En and IWSLT14 En$\rightarrow $De measured by AL and BLEU.

\begin{table*}[t]
\centering
\begin{tabular}{l c c c}
\toprule
\textbf{Hyperparameter} & \textbf{IWSLT15 En$\leftrightarrow $Vi} & \textbf{IWSLT14 De$\leftrightarrow $En} \\ \hline
encoder layers          & 6         & 6                        \\
encoder attention heads & 4         & 4                        \\
encoder embed dim       & 512       & 512                    \\
encoder ffn embed dim   & 1024      & 1024                       \\
decoder layers          & 6         & 6                       \\
decoder attention heads & 4         & 4                    \\
decoder embed dim       & 512       & 512                       \\
decoder ffn embed dim   & 1024      & 1024                      \\
dropout                 & 0.3       & 0.3                     \\
optimizer               & adam      & adam                       \\
adam-$\beta$          & (0.9, 0.98) & (0.9, 0.98)    \\
clip-norm               & 0         & 0                    \\
lr                      & 5e-4      & 5e-4                    \\
lr scheduler        & inverse sqrt  & inverse sqrt     \\
warmup-updates          & 4000      & 4000                \\
warmup-init-lr          & 1e-7      & 1e-7                    \\
weight decay            & 0.0001    & 0.0001          \\
label-smoothing         & 0.1       & 0.1                      \\
max tokens              & 16000     &  8192$\times$4  \\
\bottomrule
\end{tabular}
\caption{Hyperparameters of our experiments.}
\label{Hyper}
\end{table*}

\begin{table}[]
\centering
\begin{tabular}{p{2cm}<{\centering} p{2cm}<{\centering} p{2cm}<{\centering}} 
\toprule[1.5pt]
\multicolumn{3}{c}{\textbf{IWSLT15 En$\rightarrow $Vi}}                \\
\midrule[1pt]
\multicolumn{3}{c}{\textit{\textbf{Offline}}}     \\\hline
          & AL  & BLEU   \\
          & 22.41    &28.80   \\
\midrule[1pt]
\multicolumn{3}{c}{\textit{\textbf{Wait-$k$}}}     \\
\hline
    $k$      & AL  & BLEU   \\
    1      &3.03      &25.28  \\
    3      &4.64      &27.53  \\
    5      &6.46      &28.27  \\
    7      &8.11      &28.45  \\
    9      &9.80      &28.53  \\
\midrule[1pt]
\multicolumn{3}{c}{\textit{\textbf{Multi-path}}}     \\
\hline
    $k$      & AL  & BLEU   \\
    1      &3.16     &25.82   \\
    3      &4.69     &27.99  \\
    5      &6.42     &28.33  \\
    7      &8.17     &28.39  \\
    9      &9.82     &28.36 \\
\midrule[1pt]

\multicolumn{3}{c}{\textit{\textbf{Translation-based}}}     \\
\hline
    N/A    & AL  & BLEU   \\
    N/A      &0.61       &21.92   \\
\midrule[1pt]

\multicolumn{3}{c}{\textit{\textbf{MMA}}}     \\
\hline
    $\lambda$      & AL  & BLEU   \\
    0.4      &2.68      &27.73   \\
    0.2      &3.57      &28.47  \\
    0.1      &4.63      &28.42  \\
    0.04     &5.44      &28.33  \\
    0.02     &7.09      &28.28 \\
\midrule[1pt]

\multicolumn{3}{c}{\textit{\textbf{BS-SiMT}}}     \\
\hline
    \textbf{[$l_1, r_1$]}    & AL  & BLEU   \\
    \text{[1, 5]}    &2.00    &28.13 \\
    \text{[3, 7]}    &3.40    &28.00 \\
    \text{[5, 9]}   &5.39    &29.05 \\
    \text{[7, 11]}   &7.29    &28.86 \\
    \text{[9, 13]}   &9.07    &29.04 \\
\midrule[1pt]

\end{tabular}
\caption{Numerical results of IWSLT15 En$\rightarrow$Vi.}
\label{envi}
\end{table}

\begin{table}[]
\centering
\begin{tabular}{p{2cm}<{\centering} p{2cm}<{\centering} p{2cm}<{\centering}} 
\toprule[1.5pt]
\multicolumn{3}{c}{\textbf{IWSLT15 Vi$\rightarrow $En}}                \\
\midrule[1pt]
\multicolumn{3}{c}{\textit{\textbf{Offline}}}     \\\hline
          & AL  & BLEU   \\
          & N/A    &26.11   \\
\midrule[1pt]
\multicolumn{3}{c}{\textit{\textbf{Wait-$k$}}}     \\
\hline
    $k$      & AL  & BLEU   \\
    3      &1.49      &17.44  \\
    5      &3.28      &19.02  \\
    7      &6.75      &22.39  \\
    9      &7.91      &23.28  \\
\midrule[1pt]
\multicolumn{3}{c}{\textit{\textbf{Multi-path}}}     \\
\hline
    $k$      & AL  & BLEU   \\
    3      &1.75     &20.13  \\
    5      &4.26     &22.73  \\
    7      &6.51     &23.71  \\
    9      &8.50     &24.81 \\
\midrule[1pt]

\multicolumn{3}{c}{\textit{\textbf{Translation-based}}}     \\
\hline
    N/A    & AL  & BLEU   \\
    N/A      &3.83       &23.93   \\
\midrule[1pt]

\multicolumn{3}{c}{\textit{\textbf{MMA}}}     \\
\hline
    $\lambda$      & AL  & BLEU   \\
    0.4      &4.26      &22.08   \\
    0.2      &5.03      &23.50  \\
    0.1      &5.70     &24.15  \\
    0.05     &7.51      &24.26  \\
\midrule[1pt]

\multicolumn{3}{c}{\textit{\textbf{BS-SiMT}}}     \\
\hline
    \textbf{[$l_1, r_1$]}    & AL  & BLEU   \\
    \text{[3, 7]}    &3.90    &24.99 \\
    \text{[5, 9]}   &5.05    &25.31 \\
    \text{[7, 11]}   &6.68    &26.13 \\
    \text{[9, 13]}   &9.30    &26.68 \\
\midrule[1pt]

\end{tabular}
\caption{Numerical results of IWSLT15 Vi$\rightarrow$En.}
\label{vien}
\end{table}

\begin{table}[]
\centering
\begin{tabular}{p{2cm}<{\centering} p{2cm}<{\centering} p{2cm}<{\centering}}
\toprule[1.5pt]
\multicolumn{3}{c}{\textbf{IWSLT14 De$\rightarrow $En}}                \\
\midrule[1pt]
\multicolumn{3}{c}{\textit{\textbf{Offline}}}     \\\hline
          & AL  & BLEU   \\
          & N/A    &33   \\
\midrule[1pt]
\multicolumn{3}{c}{\textit{\textbf{Wait-$k$}}}     \\
\hline
    $k$      & AL  & BLEU   \\
    1      &0.19      &20.37  \\
    3      &1.97      &26.41  \\
    5      &3.05      &28.07  \\
    7      &4.02      &29.20  \\
    9      &6.16      &31.14  \\
    11     &8.02      &31.83  \\
\midrule[1pt]
\multicolumn{3}{c}{\textit{\textbf{Multi-path}}}     \\
\hline
    $k$      & AL  & BLEU   \\
    1      &0.74     &22.07  \\
    3      &2.53     &27.36  \\
    5      &4.43     &29.90  \\
    7      &6.07     &30.77  \\
    9      &7.93     &31.49 \\
\midrule[1pt]

\multicolumn{3}{c}{\textit{\textbf{Translation-based}}}     \\
\hline
    N/A    & AL  & BLEU   \\
    N/A      &0.2       &26.70   \\
\midrule[1pt]

\multicolumn{3}{c}{\textit{\textbf{MMA}}}     \\
\hline
    $\lambda$      & AL  & BLEU   \\
    0.4      &3.11      &24.98   \\
    0.2      &4.05      &28.00  \\
    0.1      &4.57     &28.45  \\
    0.05      &5.45      &30.03  \\
    0.01      &7.31     &20.89  \\
\midrule[1pt]

\multicolumn{3}{c}{\textit{\textbf{BS-SiMT}}}     \\
\hline
    \textbf{[$l_1, r_1$]}    & AL  & BLEU   \\
    \text{[3, 7]}    &3.26    &28.95 \\
    \text{[5, 9]}    &5.01    &30.44 \\
    \text{[7, 11]}   &7.00    &31.37 \\
    \text{[9, 13]}   &8.77    &31.96 \\
\midrule[1pt]
\end{tabular}
\caption{Numerical results of IWSLT14 De$\rightarrow$En.}
\label{deen}
\end{table}

\begin{table}[]
\centering
\begin{tabular}{p{2cm}<{\centering} p{2cm}<{\centering} p{2cm}<{\centering}}
\toprule[1.5pt]
\multicolumn{3}{c}{\textbf{IWSLT14 En$\rightarrow $De}}                \\
\midrule[1pt]
\multicolumn{3}{c}{\textit{\textbf{Offline}}}     \\\hline
          & AL  & BLEU   \\
          & 23.25    &27.18   \\
\midrule[1pt]
\multicolumn{3}{c}{\textit{\textbf{Wait-$k$}}}     \\
\hline
    $k$      & AL  & BLEU   \\
    1      &2.03      &18.54  \\
    3      &3.31      &22.30  \\
    5      &5.17      &25.45  \\
    7      &6.83      &26.01  \\
    9      &8.52      &25.64  \\
\midrule[1pt]
\multicolumn{3}{c}{\textit{\textbf{Multi-path}}}     \\
\hline
    $k$      & AL  & BLEU   \\
    3      &3.22     &23.50  \\
    5      &5.01     &25.84  \\
    7      &6.84     &26.65  \\
    9      &8.64     &26.83 \\
\midrule[1pt]

\multicolumn{3}{c}{\textit{\textbf{Translation-based}}}     \\
\hline
    N/A    & AL  & BLEU   \\
    N/A      &-2.0       &15.00   \\
\midrule[1pt]

\multicolumn{3}{c}{\textit{\textbf{MMA}}}     \\
\hline
    $\lambda$      & AL  & BLEU   \\
    0.4      &4.27      &24.06   \\
    0.2      &5.28      &24.28  \\
    0.1      &7.16     &24.33  \\
\midrule[1pt]

\multicolumn{3}{c}{\textit{\textbf{BS-SiMT}}}     \\
\hline
    \textbf{[$l_1, r_1$]}    & AL  & BLEU   \\
    \text{[3, 7]}    &4.18    &25.53 \\
    \text{[5, 9]}    &5.66    &26.73 \\
    \text{[7, 11]}   &6.56    &27.26 \\
    \text{[9, 13]}   &8.40    &27.31 \\
\midrule[1pt]
\end{tabular}
\caption{Numerical results of IWSLT14 En$\rightarrow$De.}
\label{ende}
\end{table}

\end{document}